\documentclass[sigconf]{acmart}

\AtBeginDocument{%
	\providecommand\BibTeX{{%
			\normalfont B\kern-0.5em{\scshape i\kern-0.25em b}\kern-0.8em\TeX}}}

\copyrightyear{2021}
\acmYear{2021}
\setcopyright{acmcopyright}\acmConference[MM '21]{Proceedings of the 29th ACM International Conference on Multimedia}{October 20--24, 2021}{Virtual Event, China}
\acmBooktitle{Proceedings of the 29th ACM International Conference on Multimedia (MM '21), October 20--24, 2021, Virtual Event, China}
\acmPrice{15.00}
\acmDOI{10.1145/3474085.3475655}
\acmISBN{978-1-4503-8651-7/21/10}

\acmSubmissionID{2650}

\usepackage{xcolor}
\usepackage{bbding} 
\usepackage{amsmath}
\usepackage{balance}
\usepackage{url}

\newcommand{\mypm}{$\pm$\,}

\begin{document}

	\fancyhead{} 
	
	\title{Meta-FDMixup: Cross-Domain Few-Shot Learning \\Guided by Labeled Target Data}
	
	\author[Y. Fu, Y. Fu, Y.-G. Jiang]{Yuqian Fu$^{1}$, Yanwei Fu$^{2}$, Yu-Gang Jiang$^{1\#}$}
	\affiliation{$^1$Shanghai Key Lab of Intelligent Information Processing, School of Computer Science, Fudan University 
	}
	\affiliation{$^2$School of Data Science, Fudan University
	}
	\affiliation{\{fuyq20, yanweifu, ygj\}@fudan.edu.cn
		\country{}}
	\thanks{$\#$ indicates corresponding author}
	
	\renewcommand{\shortauthors}{Fu, Fu and Jiang, et al.}

	\begin{abstract}
		A recent study~\cite{chen2019closer} finds that existing few-shot learning methods, trained on the source domain, fail to generalize to the novel target domain when a domain gap is observed. This motivates the task of Cross-Domain Few-Shot Learning (CD-FSL). In this paper, we realize that the labeled target data in CD-FSL has not been leveraged in any way to help the learning process. Thus, we advocate utilizing few labeled target data to guide the model learning. Technically, a novel \emph{meta-FDMixup} network is proposed. We tackle this problem mainly from two aspects. Firstly, to utilize the source and the newly introduced target data of two different class sets, a mixup module is re-proposed and integrated into the meta-learning mechanism. Secondly, a novel disentangle module together with a domain classifier is proposed to extract the disentangled domain-irrelevant and domain-specific features. These two modules together enable our model to narrow the domain gap thus generalizing well to the target datasets. Additionally, a detailed feasibility and pilot study is conducted to reflect the intuitive understanding of CD-FSL under our new setting. Experimental results show the effectiveness of our new setting and the proposed method. Codes and models are available at \textcolor{blue}{\url{https://github.com/lovelyqian/Meta-FDMixup}}.
		
	\end{abstract}

	\ccsdesc[500]{Computing methodologies~Computer vision}
	\ccsdesc[500]{Computing methodologies~Machine learning}

	\keywords{Cross-domain few-shot learning; Feature disentanglement; Mixup}

	\maketitle

	\section{Introduction}
	
	Few-shot Learning (FSL) aims at recognizing novel target classes with only one or few labeled examples. Many efforts have been made in tackling this problem, such as ProtoNet~\cite{snell2017prototypical}, RelationNet~\cite{sung2018learning}, MatchingNet~\cite{vinyals2016matching}, and GNN~\cite{garcia2017few}. In general, FSL assumes that the source and the target datasets belong to the same domain. Unfortunately, if a large domain gap between the source and the target datasets is observed, existing FSL methods may fail to generalize to the target datasets~\cite{chen2019closer}. This motivates the recent study of \emph{Cross-Domain Few-Shot Learning (CD-FSL)} which aims at learning robust FSL classifiers for the target datasets from an avalanche of training data in the source domain.

	\begin{figure}[t]
		\centering
		{\includegraphics[width=0.99\linewidth]{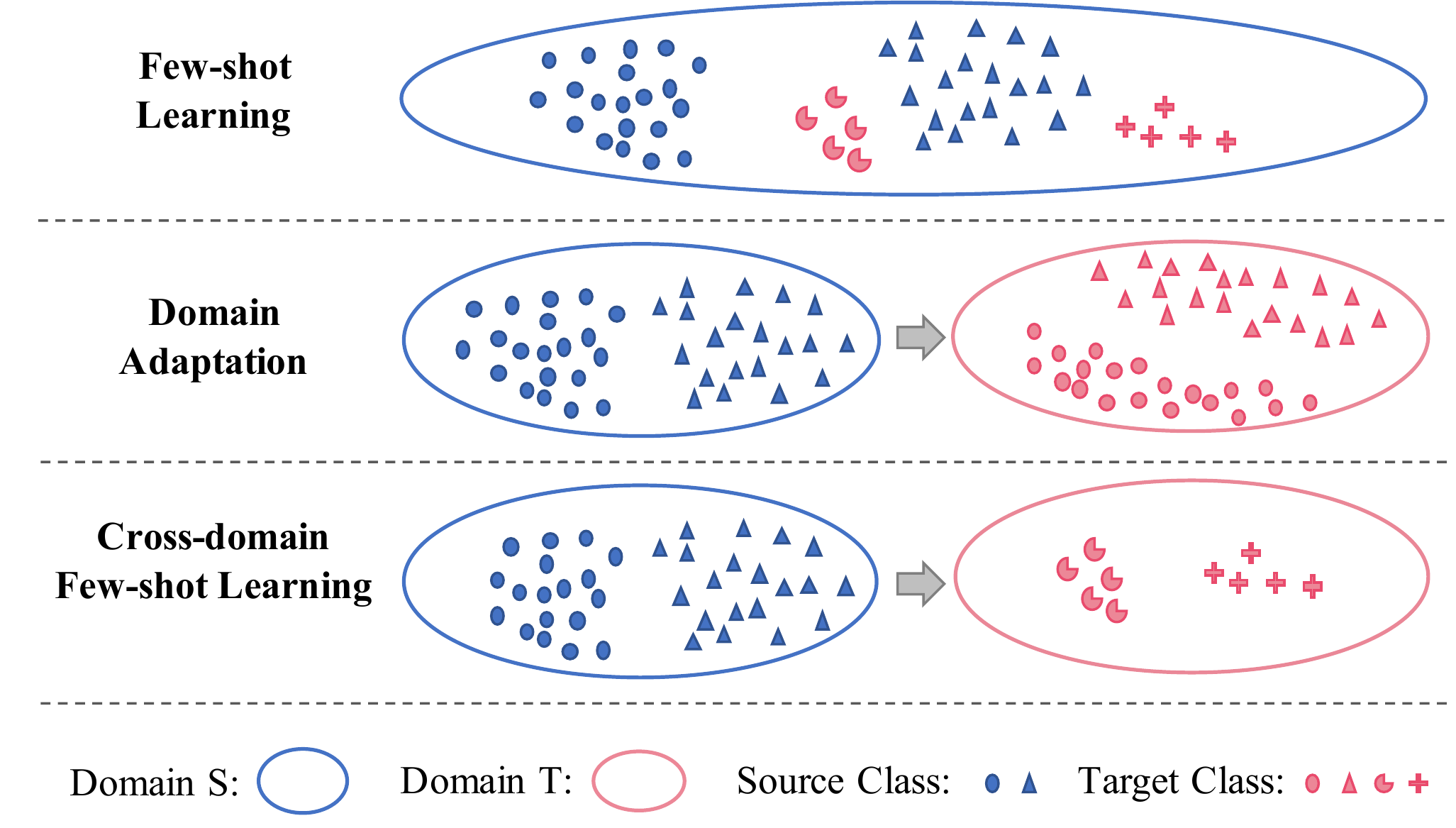}}
		\caption{A comparison of few-shot learning, domain adaptation, and cross-domain few-shot learning is illustrated.
		}
		\label{fig:CD-FSl} 
		 \vspace{-0.15in}
	\end{figure}
	
	
	A comparison of FSL, domain adaption, and CD-FSL is shown in Figure~\ref{fig:CD-FSl}. Essentially, CD-FSL is challenged in addressing the FSL and domain adaptation tasks, simultaneously. 
	There is very little work proposed for CD-FSL. 
	Recently, FWT~\cite{tseng2020cross} proposes a feature-wise transformation layer to improve the generalization ability of the model. The hyper-parameters of the transformation layers are either manually tuned in a single-source setting or learned by constructing a pseudo-seen dataset and another pseudo-unseen dataset in a multiple-source setting. In this paper, we mainly focus on the single-source setting which is more practical in real-world applications. LRP-GNN~\cite{sun2020explanation} and SB-MTL~\cite{cai2020sb} tackle this problem mainly by using the results of the explanation model and mapping the prediction scores onto a metric space, respectively. 
	Another work~\cite{phoo2020self} supposes that a large number of unlabeled target examples are available. It is also a practical setting in some cases, particularly for analyzing medical images. But it may not be an easy task in many cases to collect so much unlabeled data such as fine-grained birds and plants.

	Different from the previous works, this paper considers a more realistic scenario in CD-FSL. Particularly, in the CD-FSL task, the target classes always have a few labeled examples. We advocate utilizing these extremely few labeled target data (termed as the \emph{auxiliary dataset}) to guide the meta-learning process. 	
	This has been proposed for two reasons. Firstly, without guidance from such target data, vanilla meta-learning algorithms may fail as both classes and domains are different between the source and the target datasets. Secondly, in practice, it may be much easier to label a few images than collect a large number of unlabeled images.
	Notably, all the images contained in the auxiliary dataset will not appear during the testing stage, thus we do not violate the basic CD-FSL setting.
	
	Detailed pilot and feasibility studies are conducted under our new setting including 1) which training stage should the auxiliary data be used; 2) how the number of auxiliary data will affect the performance of our model. Our experimental results suggest the most appropriate training strategy under this setting and validate the soundness of our setting. A significant performance gain can be brought by introducing a small number of auxiliary target data.

	Technically, we tackle this problem from two aspects. Firstly, a \emph{mixup module} is re-proposed to make the best use of instances from the source and the newly introduced auxiliary datasets/domains. 
	Mixing the images of two domains augments the auxiliary dataset to a large extent.  Secondly, a \emph{disentangle module} is proposed to decouple the image features into the disentangled domain-irrelevant and domain-specific features. The former ones are used to conduct the few-shot learning across the domains, while the latter ones contain significant inductive bias learned from the domains.
	
	Formally,  a novel network termed as \emph{\textbf{meta-FDMixup} (\textbf{meta-}learning based \textbf{F}eature-\textbf{D}istangled \textbf{Mixup}}) is proposed in this paper. Specifically, for each training step, we first sample two episodes from the source and the auxiliary datasets, respectively. We then mix up the query images of the source episode with that of the auxiliary episode at a certain ratio. After that, our feature extractor and disentangle module learn to extract domain-irrelevant and domain-specific features for the source support set, the auxiliary support set, and the mixed query set, individually. 
	Then, FSL classification tasks and domain classification tasks are performed. For FSL classification tasks, our FSL classifier is forced to classify the mixed query images into the class sets of the source support images and the auxiliary support images, given the domain-irrelevant features. For domain classification tasks, a domain classifier that aims at predicting the domain categories correctly only by domain-specific features is proposed. These two tasks are performed simultaneously.
	
	Our contributions are summarized as follows. 
	1) For the first time, we advocate introducing extremely few labeled target data to guide the meta-learning process for the CD-FSL task.
	2) Detailed pilot and feasibility studies are conducted for this new setting which reflect our intuitive understanding of CD-FSL, with the hope of igniting future research.
	3) Technically, a novel \emph{meta-FDMixup} network is proposed which utilizes the source and the newly introduced auxiliary data well by mixing the images and learns to disentangle the images features into the domain-irrelevant and the domain-specific ones, improving the model performance on both the source and the target domains.
	4) The extensive experimental results validate the soundness of our new setting and the effectiveness of our proposed method.

	\section{Related Work}
	\noindent 
	\textbf{Few-shot Learning.}
	Many efforts have been done to tackle the FSL problem. Existing few-shot learning methods mainly focus on model initialization~\cite{finn2017model,rusu2018meta}, metric-learning~\cite{snell2017prototypical, vinyals2016matching, sung2018learning, garcia2017few}, and data augmentation~\cite{chen2019image,li2020adversarial}. More recent work includes cross-attention-based CAN~\cite{hou2019cross},  set-to-set function based FEAT~\cite{ye2020few}, episodes relation based MELR~\cite{feimelr} and so on.
	One common point of the above methods is that they sample the source and the target images from the same dataset, intrinsically belonging to the same domain. Research~\cite{chen2019closer} points out that existing metric-learning based few-shot methods fail to generalize to the novel target classes when the source and the target data are sampled from different datasets. Thus, in this paper, we mainly stick to the metric-learning based methods and try to improve their performance under the cross-domain setting.

	\noindent 
	\textbf{Domain Adaptation.}
	Domain adaption aims at transferring knowledge from the source domain to the target domain which has the same class set while different data distribution from that of the source domain. 
	Deep learning based methods can be roughly organized into three types: discrepancy based methods~\cite{tzeng2014deep, rozantsev2018beyond, kang2019contrastive}, adversarial based methods~\cite{tzeng2015simultaneous, ganin2016domain, bousmalis2017unsupervised}, and reconstruction based methods~\cite{bousmalis2016domain,ghifary2016deep}.  Though our idea of extracting the domain-irrelevant features keeps the same with these domain adaptation methods, we are still different from them mainly in two aspects. Firstly, the class sets of the source and the target domains are the same in domain adaptation while disjoint in CD-FSL. Another point is that sufficient labeled or unlabeled target data can be used in the domain adaption task in general, while we have to classify the novel classes with few labeled data.

	\noindent 
	\textbf{Cross-domain Few-shot Learning.}
	Greatly promoted by the research~\cite{chen2019closer}, the task of cross-domain few-shot learning is formally proposed in FWT~\cite{tseng2020cross}. Then, a new broader study together with a new benchmark is contributed by BSCD-FSL~\cite{guo2020broader}. Methods using only source data including FWT~\cite{tseng2020cross}, LRP-GNN~\cite{sun2020explanation} and SB-MTL~\cite{cai2020sb}. STARTUP~\cite{phoo2020self} relaxes the setting and assumes that the model has access to many unlabeled target data during training. Particularly, this paper resorts to a novel setting in that extremely few labeled target images are available. Comparing to the former strictly defined setting, a significant performance gain can be achieved with only a small number of labeled data. Comparing to the latter one, collecting extremely few labeled samples, e.g. 5 images per class, maybe easier and more practical than constructing a large amount of unlabeled data, especially in the case of fine-grained classes.

	\noindent 
	\textbf{Data Augmentation.}
	Recently, an advanced data augmentation termed as mixup~\cite{zhang2017mixup} is proposed to improve the robustness of the network. After that, several variants of mixup are proposed, including CutMix~\cite{yun2019cutmix}, Manifold Mixup~\cite{verma2019manifold}, AugMix~\cite{hendrycks2019augmix}, PuzzleMix~\cite{kim2020puzzle} and so on. These mixup methods are designed for the classical classification task, which means that the mixed images come from the same class set and the model is trained with the standard cross-entropy loss. 
	More related work to us is Xmixup~\cite{li2020xmixup} proposed for cross-domain transfer learning. Given a disjoint source and target dataset, XMixup assumes the source has more classes than the target. It first selects a unique source class for each target class, then applies mixup onto the one-to-one pairs. 
	However, Xmixup still can not handle the few-shot learning problem.

	\begin{figure*}[h!]
		\centering
		{\includegraphics[width=0.99\linewidth]{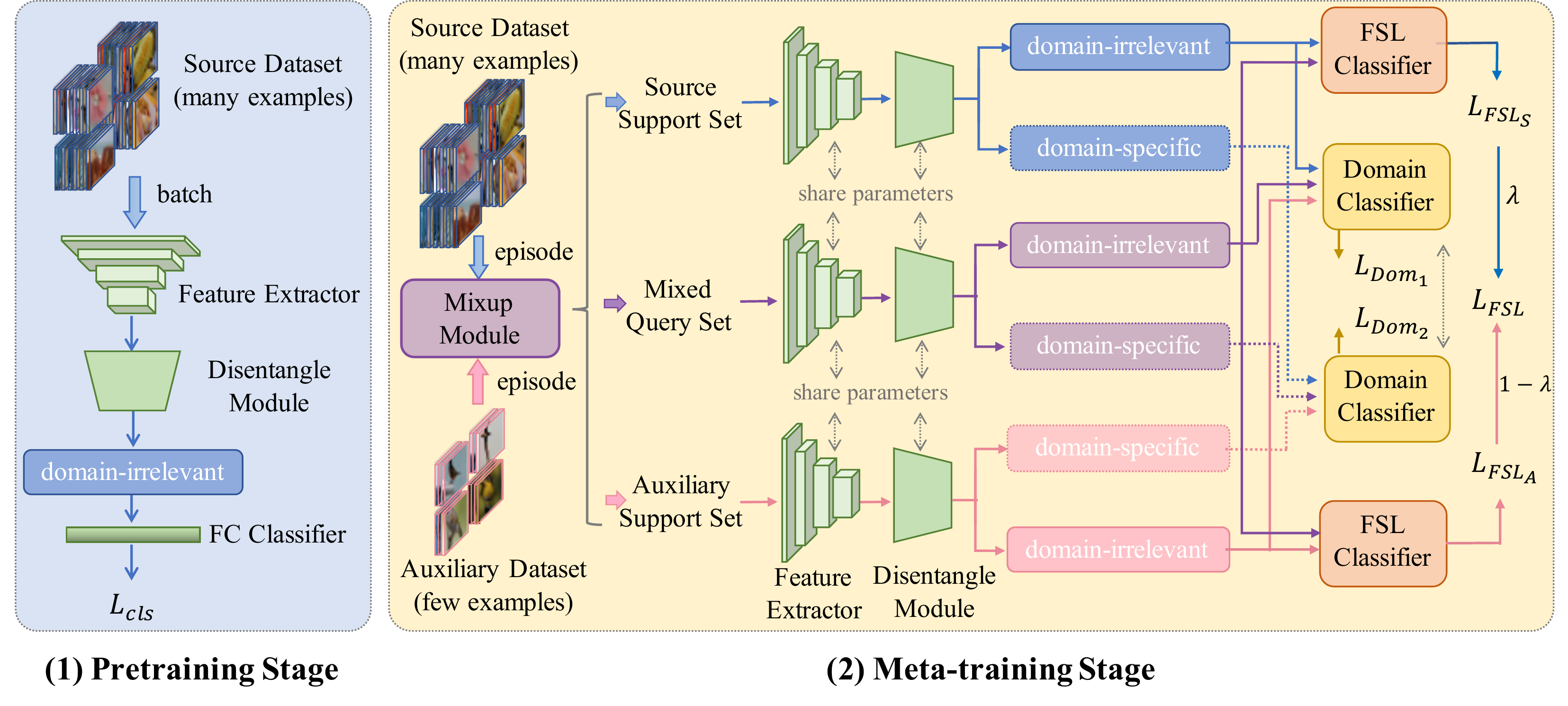}}
		\caption{Overview of our framework. It mainly contains two stages. During the pretraining stage, we pretrain the feature extractor,  the disentangle module, and the FC classifier with the supervised classification tasks. During  the meta-training stage, we finetune the whole network with episode-based few-shot classification tasks in the meta-learning paradigm. To make use of the source and the newly introduced auxiliary datasets, a mixup module is re-proposed. Furthermore, to narrow the domain gap, we learn to disentangle the features by introducing a  disentangle module and a domain classifier.}
		\label{fig:framework} 
	\end{figure*}

	\section{Method}\label{sec:method}

	\subsection{Problem Formulation}
	Given the source dataset $D_{S}=\left\{ \mathbf{I}_{j,}y_{j}\right\} _{j=1}^{n_{S}}$, of sufficient labeled images $y_{j}\in\mathcal{C}_{S}$, and a target data $D_{T}=\left\{ \mathbf{I}_{j,}y_{j}\right\} _{j=1}^{n_{T}}$
	with few labeled images, $y_{j}\in\mathcal{C}_{T}$. We have the disjointed
	source and target classes, \emph{i.e.}, $C_{S}\cap C_{T}=\emptyset$.
	Different from the standard few-shot learning where the $D_{S}$ and
	the $D_{T}$ are sampled from the same domain, the data distributions
	of the $D_{S}$ and the $D_{T}$ are different from each other in
	CD-FSL. The goal of our model is to learn the knowledge from the source
	dataset $D_{S}$ and transfer it to the target dataset $D_{T}$.

	In the standard CD-FSL setting proposed by FWT~\cite{tseng2020cross}, the source dataset $D_{S}$ is further split into a source \emph{base} $D_{S_{b}}$, a source \emph{eval} $D_{S_{e}}$, and a source \emph{novel} $D_{S_{n}}$. Similarly, the target dataset $D_{T}$ is split into a target \emph{base} $D_{T_{b}}$, a target \emph{eval} $D_{T_{e}}$, and a target \emph{novel} $D_{T_{n}}$. The class sets of the base, the eval, and the novel datasets are disjointed from each other, again. During the experiments, the $D_{S_{b}}$, $D_{S_{e}}$, and $D_{T_{n}}$ serve as the training set, validation set, and testing set, respectively.

	As for our new setting, we randomly sample $num_{target}$ labeled images per class from the target base dataset $D_{T_{b}}$. The selected target images together constitute our auxiliary dataset $D_{aux}$. Notably and most importantly, since the auxiliary dataset $D_{aux}$ comes from the target base dataset $D_{T_{b}}$ which is disjointed from the target testing set $C_{T_{n}}$, thus our setting does not violate the basic setting of CD-FSL.

	\noindent\textbf{N-way-K-shot Problem.}
	To mimic the testing phase, meta-learning adopts an episode-based training mechanism. Each episode contains a support set $S=\left\{x_i, y_i\right\}_{i=1}^{N\times K}$  and a query set $Q=\left\{x_i, y_i\right\}_{i=1}^{N\times M}$. More specially, given a dataset, we first randomly sample $N$ classes, then for each selected class, K labeled images and another $M$ images are randomly sampled to constitute the support set and the query set, respectively. Our model learns to classify the labels of the query images according to the support set.

	\subsection{Meta-FDMixup Network}\label{sec:model}
	The illustration of our meta-FDMixup network is shown in Figure~\ref{fig:framework}.
	Meta-FDMixup mainly consists of a feature extractor $f$, a disentangle module $h$, a FC classifier $g_{cls}$, a few-shot learning (FSL) classifier $g_{fsl}$, and a domain classifier $g_{dom}$. The according parameters are denoted as $\theta_{f}$, $\theta_{h}$, $\theta_{cls}$,
	$\theta_{fsl}$, and $\theta_{dom}$, respectively. The feature extractor
	$f$ extracts a 1-D feature representation $F$ for the input
	image. Principally, any image classification network can be introduced
	as the encoder. The disentangle module $h$ further decouples the
	extracted 1-D feature $F$ into a \emph{domain-irrelevant feature}
	and a \emph{domain-specific feature}. We use the $H1$ and $H2$ to
	denote the domain-irrelevant and the domain-specific features, respectively.
	The FC classifier $g_{cls}$ means a single fully connected layer
	that maps the extracted image representation to the scores on the
	source base classes $C_{S_{b}}$. The FSL classification $g_{fsl}$ takes a support set $S$ and a query set $Q$ as the input and is responsible for classifying the images in the query set according to the given support set. The domain classifier $g_{dom}$ learns to classify the input
	image into the source domain or the target domain. It works
	as the discriminator to supervise the learning process of the disentangle
	module.

	A two-step training strategy is adopted to train the model.
	As for the pretraining stage, the batch data sampled from the source
	base dataset $D_{S_{b}}$ is fed into the feature extractor $f$ and
	the disentangle module $h$ successively. Then, the extracted domain-irrelevant feature $H1$ is fed into the FC classifier $g_{cls}$ to obtain its predictions on the source base classes $C_{S_{b}}$. The standard cross-entropy classification loss is used to train the network. During this process, the $\theta_{f}$, the $\theta_{h}$, and the $\theta_{cls}$ are optimized.

	As for the meta-training stage, we train the whole network including the feature extractor $f$, the disentangle module $h$, the FSL classifier $g_{fsl}$, and the domain classifier $g_{dom}$ with the episode data under the meta-learning paradigm. For each training iteration, two episodes are randomly sampled from the source base dataset $D_{S_b}$ and the auxiliary dataset $D_{aux}$, respectively. In order to make the best use of the initial source data and the newly introduced auxiliary data, a simple yet effective \emph{mixup module} is proposed which collaborates with the meta-learning mechanism. More specifically, given a source episode $E_{sou}$ and an auxiliary episode $E_{aux}$, we mix the images contained in the source query set $Q_{sou}$ with that in the auxiliary query set $Q_{aux}$ through linear combination at a ratio of $\lambda$.
	\begin{equation}
		Q_{mix} = \lambda Q_{sou} + (1-\lambda) Q_{aux},
	\end{equation}
	where the $Q_{mix}$ denotes the mixed query set, and the ratio $\lambda \sim Beta(\alpha, \alpha),\alpha \in (0,\infty)$ keeps the same as the mixup~\cite{zhang2017mixup}.
	
\begin{figure}[h]
		\centering
		{\includegraphics[width=0.99\linewidth]{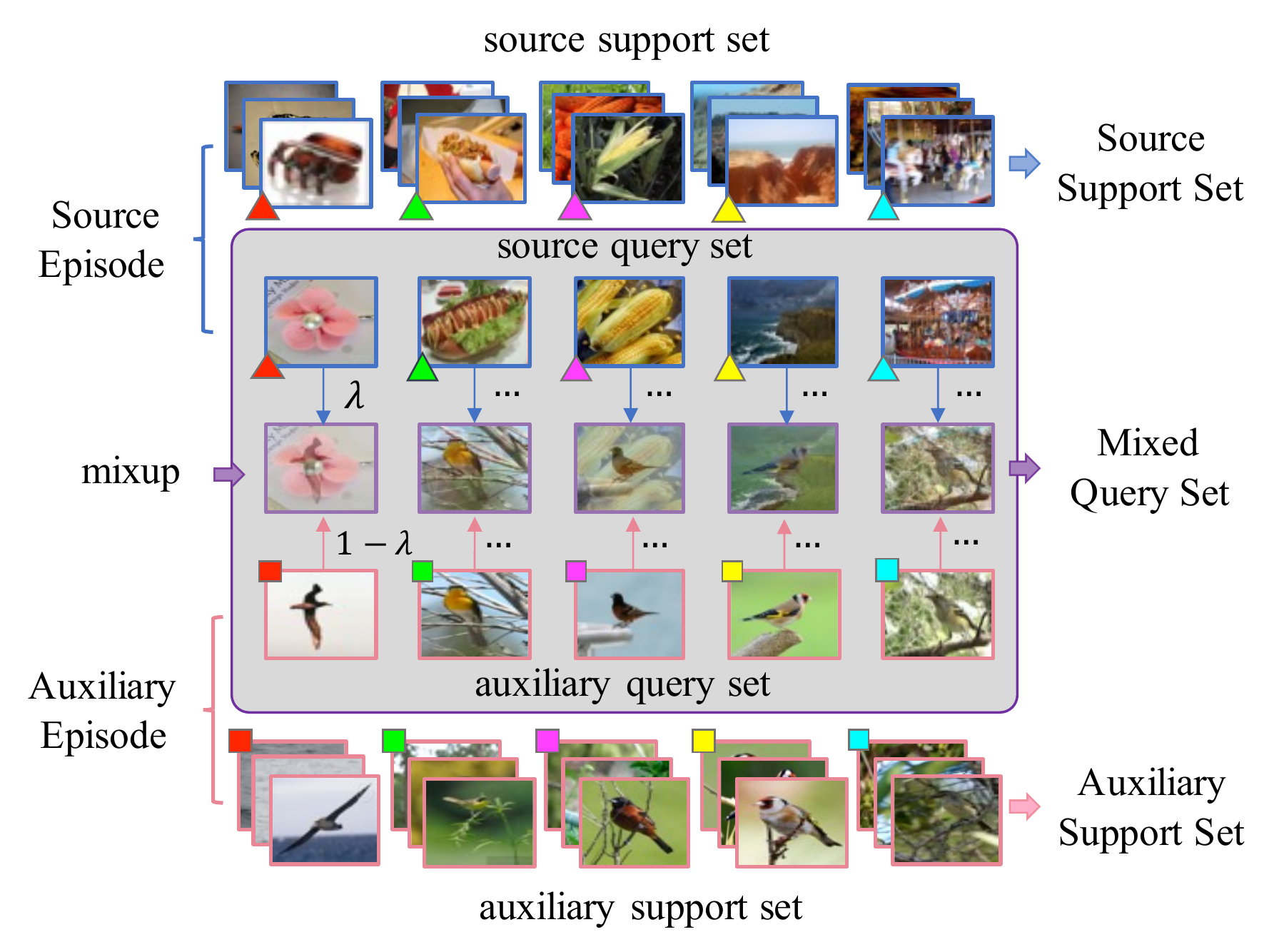}}
		\caption{Illustration of our mixup module. We mix the query sets of the source and the auxiliary episodes at a ratio of $\lambda$ while keep the support sets of them unchanged. }
		\label{fig:mixup} 
	\end{figure}

	Considering the FSL classification task requires a reliable support set, we keep the support sets of both the source episode and the auxiliary episode unchanged. The detailed illustration of our mixup module is shown in Figure~\ref{fig:mixup}. Formally, taking the source episode $E_{sou}$ and the auxiliary episode $E_{aux}$ as the input, our mixup module generates a source support set $S_{sou}$, an auxiliary support set $S_{aux}$, a mixed query set $Q_{mix}$, and the mixing ratio $\lambda$.

	After that, we feed the source support set $S_{sou}$, the auxiliary support set $S_{aux}$, and the mixed query set $Q_{mix}$ into the feature extractor $f$ and the disentangle module $h$ successively to obtain their domain-irrelevant features and domain-specific features. Take the mixed query set $Q_{mix}$ as an example, for each image contained, the feature extractor $f$ extracts a 1-D feature representation $F_{Q_{mix}}$. Then, the disentangle module further generates the domain-irrelevant feature $H1_{Q_{mix}}$ and the domain-specific feature $H2_{Q_{mix}}$ for the mixed query set $Q_{mix}$.  In the same way, for the source and the auxiliary support sets, the domain-irrelevant features  $H1_{S_{sou}}$, $H1_{S_{aux}}$ and the domain-specific features $H2_{S_{sou}}$, $H2_{S_{aux}}$ are generated.

\begin{figure}[t]
		\centering
		{\includegraphics[width=0.99\linewidth]{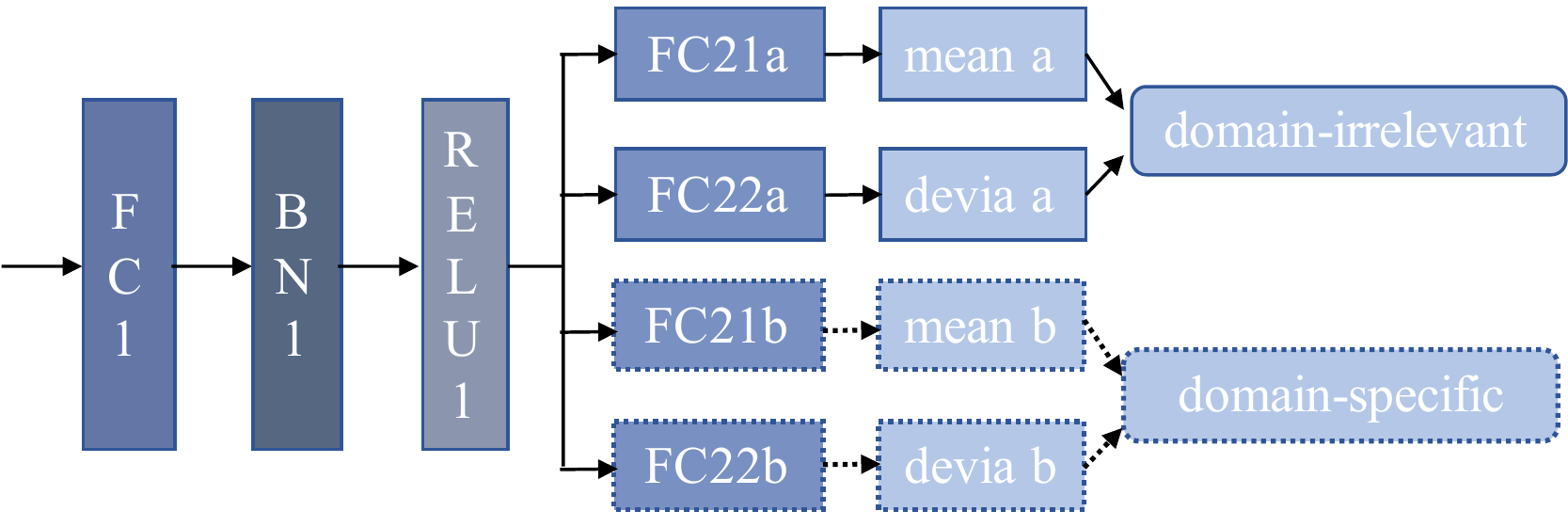}}
		\caption{Architecture of our disentangle module.}
		\label{fig:encoder} 
	\end{figure}

	The detailed illustration of the disentangled module $h$ is shown in Figure~\ref{fig:encoder}. It is mainly composed of a batch normalization layer, a relu activation layer, and several fully connected layers. More specifically, ``FC1'' is used to extract the general representation, while ``FC21a'' and ``FC22a'' are responsible for extracting the domain-irrelevant features -- mean a and devia a. Correspondingly, the ``FC21b'' and ``FC22b'' are designed for extracting the domain-specific features. We highlight that our disentangle module is inspired but very different from VAE (Variation Auto-Encoder). Particularly, VAE only encodes one latent feature, while the key idea of our module lies in best learning to disentangle the domain-specific and domain-irrelevant features thus alleviating the domain shift problem in CD-FSL.

	In the end, a multi-task learning mechanism composed of FSL classification tasks and domain classification tasks is proposed to optimize our model. For FSL classification tasks, the domain-irrelevant features $H1_{S_{sou}}$, $H1_{S_{aux}}$, and $H1_{Q_{mix}}$ are used for classification. Recall that our query set is a mixture of the source query set and the auxiliary query set in the ratio $\lambda$, thus in our proposed meta-FDMixup method, the ratio $\lambda$ is considered as the confidence score of the mixed image belonging to its initial ground truth. More specifically, our FSL classifier is asked to classify the mixed query set into the class set of the source support set according to their features $H1_{Q_{mix}}$ and $H1_{S_{sou}}$ resulting in the FSL classification loss $\lambda L_{FSL_S}$. Similarly, the confidence score for classifying the mixed query set into the class set of auxiliary support set is $1-\lambda$, resulting in another FSL classification loss $(1-\lambda) L_{FSL_A}$.  The $L_{FSL_S}$ and $L_{FSL_A}$ are generated by the FSL classifier. The FSL loss function is defined as follows, 
	\begin{equation}
		\mathcal{L}_{FSL}=\lambda\cdot\mathcal{L}_{FSL_{S}}+(1-\lambda)\mathcal{L}_{FSL_{A}}.
	\end{equation}

	As for domain classification tasks, we hope the domain category can be easily identified given the domain-specific features $H2$ resulting in domain classification loss $L_{dom_2}$, while the domain-irrelevant features $H1$ are expected to confuse the domain classifier resulting in domain-classification loss $L_{dom_1}$. Our domain classifier contains a fully connected layer that maps the input feature into the scores on two domains.

	Formally, let 1 and 0 denote the category of source and target, respectively. We define the ground truth for the domain-specific features $H2_{S_{sou}}$, $H2_{S_{aux}}$, $H2_{Q_{mix}}$ as $y2_{S_{sou}} \in \mathcal{R}^{NK}$, $y2_{S_{aux}} \in \mathcal{R}^{NK}$, $y2_{Q_{mix1}} \in \mathcal{R}^{NM}$,  and $y2_{Q_{mix2}} \in \mathcal{R}^{NM}$, respectively. All the values in $y2_{S_{sou}}$ and $y2_{Q_{mix1}}$ are set as 1, all the values in $y2_{S_{aux}}$ and $y2_{Q_{mix2}}$ are set as 0.

	\begin{gather}
		\mathcal{L}_{dom_{2}}=\frac{1}{3}\sum\left[\mathrm{CE}\left(g_{dom}(H2_{S_{sou}}),y2_{S_{sou}}\right)\right.\\
		+\,\mathrm{CE}\left(g_{dom}(H2_{S_{aux}}),y2_{S_{aux}}\right)\\
		+\,\lambda\cdot\mathrm{CE}\left(g_{dom}(H2_{Q_{mix}}),y2_{S_{mix1}}\right)\\
		+\left.(1-\lambda)\cdot\mathrm{CE}\left(g_{dom}(H2_{Q_{mix}}),y2_{S_{mix2}}\right)\right]
	\end{gather}

	Similarly, we define the corresponding ground truth for the domain-irrelevant features  $H1_{S_{sou}}$, $H1_{S_{aux}}$, $H1_{Q_{mix}}$ as $y1_{S_{sou}} \in \mathcal{R}^{NK\times 2}$, $y1_{S_{aux}} \in \mathcal{R}^{NK\times 2}$, $y1_{Q_{mix}} \in \mathcal{R}^{NM \times 2}$. All the lines in $y1_{S_{sou}}$, $y1_{S_{aux}}$, and $y1_{Q_{mix}}$ are set to $[0.5, 0.5]$.

	\begin{gather}
		\mathcal{L}_{dom_{1}} =\frac{1}{3}\sum\left[\mathrm{KL}\left(g_{dom}(H1_{S_{sou}}),y1_{S_{sou}}\right)\right.\\
		+\,\mathrm{KL}\left(g_{dom}(H1_{S_{aux}}),y1_{S_{aux}}\right)\\
		+\,\left.\mathrm{KL}\left(g_{dom}(H1_{Q_{mix}}),y1_{Q_{mix}}\right)\right]
	\end{gather}

	The $CE()$ and the $KL()$ denotes the cross-entropy loss and the Kullback-Leibler divergence loss, respectively. These sub-tasks are performed in parallel in an end-to-end way. Our final loss function is defined as,
	
	\begin{equation}
		\mathcal{L}=\mathcal{L}_{FSL}+\mathcal{L}_{dom_{2}}+\mathcal{L}_{dom_{1}}
	\end{equation}

	\begin{table*}[t!]
		\centering
		\begin{tabular}{llllllll}
			\hline
			5-way-5-shot & $D_{aux}$ & stage & CUB & Cars & Places & Plantae \\
			\hline
			s-base  & -  & /  &  62.25 \mypm 0.65   &    44.28 \mypm 0.63   &   70.84 \mypm 0.65   & 52.53 \mypm 0.59 \\
			\hline
			a-base & \Checkmark & P1+2 & 60.26   \mypm 0.71  & 40.43   \mypm 0.60  &  56.86   \mypm 0.67  &  53.88   \mypm 0.70 \\
			& \Checkmark & P2    & \textbf{64.98   \mypm 0.67}  & \textbf{48.53   \mypm 0.64}  & \textbf{67.96   \mypm 0.68}  & \textbf{60.21 \mypm 0.69} \\
			\hline
			m-base & \Checkmark &  P1+2 & 77.07   \mypm 0.65  & 62.98   \mypm 0.69  & \textcolor{blue}{76.08   \mypm 0.62}  & 64.33   \mypm 0.69 \\
			& \Checkmark   & P2    &  \textbf{78.08   \mypm 0.60}  & \textbf{63.27   \mypm 0.70}  & 75.90   \mypm 0.67  &  \textbf{66.69   \mypm 0.68} \\
			\hline
		\end{tabular}
		\caption{Pilot study of different training strategies. The accuracy (\%) is reported with GNN as the few-shot classifier. We compare our strategy (P2) with the two-step strategy (P1+2). }
		\label{tab:2}
	\end{table*}

	\section{Experiments}
	
	\noindent\textbf{Networks Components.} To make a fair comparison with the FWT~\cite{tseng2020cross}, we follow the basic setup used in their work. For the feature encoder, Resnet-10~\cite{he2016deep} is selected. For the FSL classifier, the GNN~\cite{garcia2017few} is adopted. For the disentangle module, the input dim and the output dim of the FC1 are set as 512 and 256, respectively. All the FC21a, FC22a, FC21b, FC22b map the 256-D feature into a 64-D feature. The domain classifier further maps the 64-D feature into a 2-D vector.
	
	\noindent\textbf{Datasets.} 
	Mini-Imagenet~\cite{ravi2016optimization} serves as the source dataset $D_{S}$, and another four datasets including CUB~\cite{wah2011caltech}, Cars~\cite{krause20133d}, Places~\cite{zhou2017places}, and Plantae~\cite{van2018inaturalist} server as the target dataset $D_{T}$, respectively. 
	
	We follow the same \emph{base}/\emph{eval}/\emph{novel} splits provided by FWT~\cite{tseng2020cross}. However, in our setting, a small part of the target base set $D_{T_{b}}$ is used at the training stage.

	\noindent\textbf{Implementation Details.} The $num_{target}$ is set as 5 unless otherwise specified. The $\alpha$ is set as 1. Whether for which training stage, we train the network for 400 epochs. Notably, for miniImagenet, the model which performs best on the validation set $D_{S_e}$ is saved for testing; For all the target datasets, considering the domain gap, we uniformly use the last model for testing. The Adam with an initial learning rate of 0.001 is used as the optimizer. We conduct experiments on the 5-way-K-shot setting. The average accuracy of 1000 episodes is reported as the final result.

	\noindent 
	\textbf{Baselines and Competitors.}
	Several baselines and competitors are compared as follows. (1) To show the performance of the existing FSL methods on this CD-FSL setting, several flagship FSL methods including MatchingNet~\cite{vinyals2016matching}, RelationNet~\cite{sung2018learning}, and GNN~\cite{garcia2017few} are introduced. For these methods, no auxiliary images are given. We unify the feature extractor as ResNet-10 for them following FWT~\cite{tseng2020cross}.  (2) The GNN is adopted as the FSL classifier for all the other baselines and CD-FSL methods due to its good performance in the CD-FSL setting. Based on the ResNet-10 (feature extractor) and the GNN (FSL classifier), we build several baselines to show the effectiveness of the introduced auxiliary data. For ``s-base", we meta-train the network with the mini-Imagenet lonely. Thus, it equals the original GNN. Correspondingly, for ``a-base", the network is meta-trained with the auxiliary target data along. For our strong baseline ``m-base", the images of the mini-Imagenet and the auxiliary dataset are merged to train the model.
	(3) As for CD-FSL competitors, since the task is recently proposed, we only study FWT~\cite{tseng2020cross}. A novel feature-wise transformation layer is proposed in FWT which learns the \emph{scale} and \emph{shift} hyper-parameters of the batch normalization layer. 
	In a single-source setting, the hyper-parameters are manually defined (FT) according to the empirical results. 
	We use the hyper-parameters provided by FWT and integrate this method into our strong baseline ``m-base" resulting in the competitor ``m-base-FT".
	Notably, only the mini-Imagenet is used during the first pretraining stage for all the methods, which will be explained in the following subsection.

	
	\subsection{Pilot \& Feasibility Study}
	Considering there are two training stages, we first conduct a pilot study on which training stage should the auxiliary data be used. In addition, to validate the practicability of the proposed setting, we conduct a detailed feasibility study on how the number of auxiliary images affects the performance of the model.

	\noindent\textbf{In which training stage should the auxiliary images be used?}
	As we stated in the sec~\ref{sec:method}, there are two training stages: 1) pretraining the feature encoder; 2) meta-training the whole model. We compare the two-stage strategy (denoted as ``P1+2") which uses the auxiliary images to train the model in both stages with ``P2" which only uses the auxiliary data in the second meta-training stage.
	We conduct experiments on two different baselines, including ``a-base" and ``m-base". The results of the ``s-base" are taken as the baseline.

	The results under the 5-way-5-shot setting are shown in Table~\ref{tab:2}. We notice that no matter which method is used, the ``P2" strategy outperforms the ``P1+2". The only exception is marked in blue font in which ``P1+2" shows a negligible advantage over ``P2". This phenomenon indicates that the feature extractor trained with the mini-Imagenet has a stronger ability to extract image representations. Using the auxiliary data in the pretraining stage may cause some noises thus damaging the performance of the whole model.

\begin{figure}
		\centering
		{\includegraphics[width=0.99\linewidth]{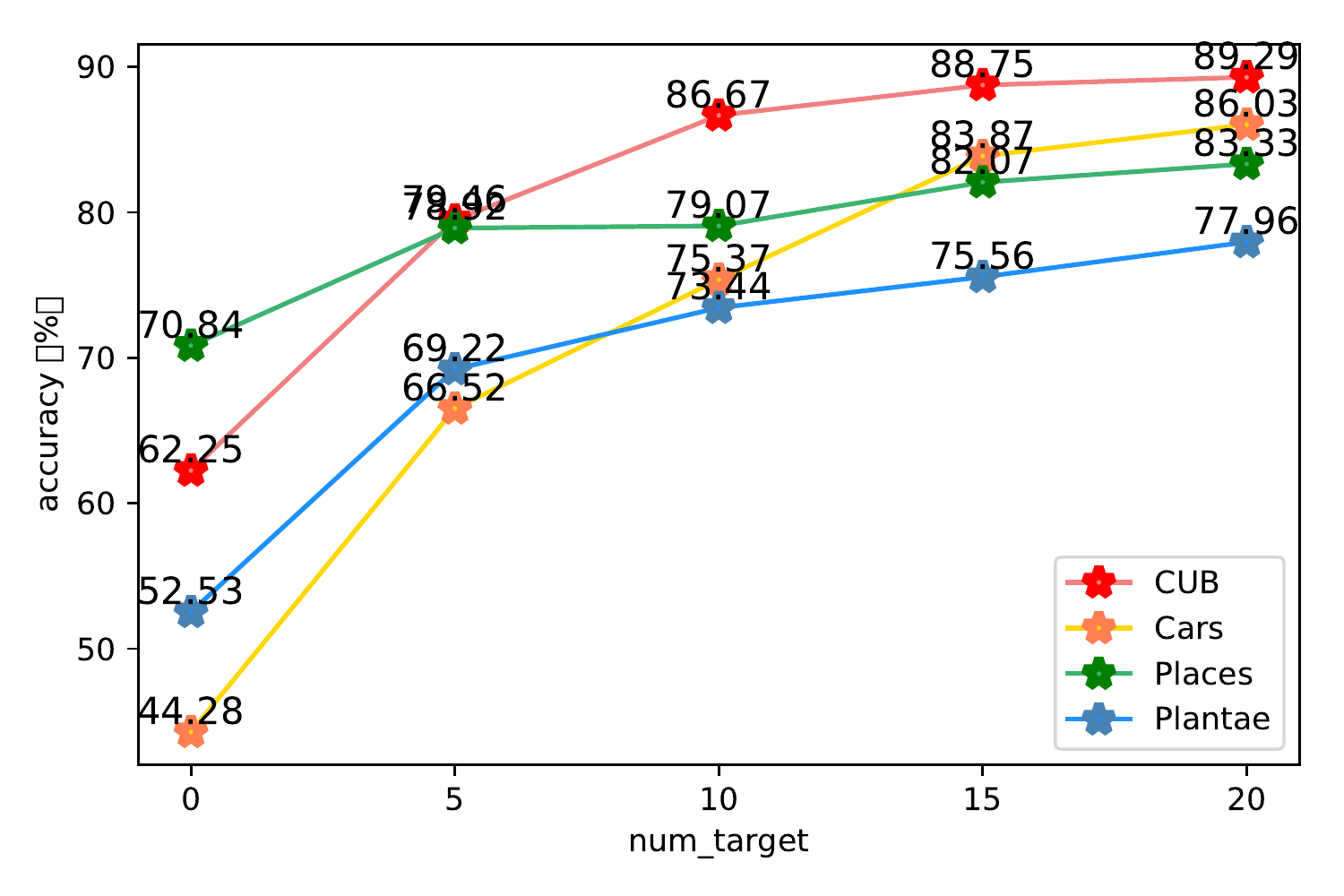}}
		\vspace{-0.15in}
		\caption{Feasibility study of our new setting. The accuracy (\%) of our meta-FDMixup method under the 5-way-5-shot setting is reported. We set the $num_{target}$ as 5, 10, 15, and 20.}
		\label{fig:study} 
		\vspace{-0.15in}
	\end{figure}

	\noindent\textbf{Is this setting feasible and how many examples should be provided?} To answer this question, we set the $num_{target}$ as 5, 10, 15, and 20 to study how the performance will be affected as the number of labeled images increases. The results of our meta-FDMixup method are reported in Figure~\ref{fig:study}.
	Take CUB as an example, 17.21\%, 24.42\%, 26.50\%, 27.04\% performance gain is achieved with $num_{target}$ equals to 5, 10, 15 and 20, respectively.
	These results validate the feasibility and soundness of our new setting.
	
	As for the most appropriate $num_{target}$, generally, the performance of the model is positively correlated with the number of labeled examples which keeps consistent with our common sense. However, we also observe that as the number of labeled examples increases, the performance improvement it brings slows down.  This can be intuitively understood by the \emph{law of diminishing marginal utility}. Take CUB as an example again, we denote 5 labeled images as a unit, the accuracy is improved by 17.21\%, 7.21\%, 2.08\%, 0.54\% by introducing the 1-st, 2-st, 3-st, and the 4-st unit, respectively. 
	To trade off the performance and the labeling cost, we set $num_{target}$ as 5 in all of our experiments.

	\subsection{Main Results}

	\begin{table*}
		\centering
		\begin{tabular}{lllllll}
			\hline
			5-way & 1-shot & $D_{aux}$ &  CUB &  Cars & Places & Plantae\\
			\hline
			FSL 
			& MatchingNet~\cite{vinyals2016matching} & - &  35.89 \mypm 0.51 & 30.77 \mypm 0.47 & 49.86 \mypm 0.79 & 32.70 \mypm 0.60  \\
			
			& RelationNet~\cite{sung2018learning} & - & 42.44 \mypm 0.77 &  29.11 \mypm 0.60 & 48.64 \mypm 0.85 &  33.17 \mypm 0.64 \\
			
			& GNN~\cite{garcia2017few} &  - & 45.69 \mypm 0.68 & 31.79 \mypm 0.51 & 53.10 \mypm 0.80 & 35.60 \mypm 0.56  \\
			
			\hline
			Baselines 
			& s-base & - &  45.69 \mypm 0.68 & 31.79 \mypm 0.51     & 53.10 \mypm 0.80 & 35.60 \mypm 0.56 \\ 
			& a-base & \Checkmark    & 50.28 \mypm 0.77 &  37.86 \mypm 0.63  &  51.09 \mypm 0.79 &  44.25 \mypm 0.74  \\
			
			& m-base  & \Checkmark & 57.65 \mypm 0.80 & 46.03 \mypm 0.72             &  55.70 \mypm 0.79 & 48.25 \mypm 0.74  \\
			\hline
			
			CD-FSL 
			& m-base-FT~\cite{tseng2020cross} &  \Checkmark &   61.16  \mypm  0.81 & 49.01  \mypm  0.76  & 57.89  \mypm  0.82  &  50.49  \mypm  0.81  \\
			& meta-FDMixup (ours) & \Checkmark &  \textbf{63.24 \mypm  0.82}  & \textbf{51.31 \mypm  0.83 }& \textbf{58.22 \mypm  0.82} & \textbf{51.03 \mypm  0.81} \\
			\hline
			\hline
			5-way & 5-shot & $D_{aux}$ & CUB & Cars & Places & Plantae\\
			\hline
			FSL 
			& MatchingNet~\cite{vinyals2016matching} &- & 51.37 \mypm 0.77 & 38.99 \mypm 0.64 & 63.16 \mypm 0.77  & 46.53 \mypm 0.68 \\
			
			& RelationNet~\cite{sung2018learning} & - & 57.77 \mypm 0.69 & 37.33 \mypm 0.68 & 63.32 \mypm 0.76 & 44.00 \mypm 0.60 \\
			
			& GNN~\cite{garcia2017few} &  - & 62.25 \mypm 0.65 &  44.28 \mypm 0.63 & 70.84 \mypm 0.65 & 52.53 \mypm 0.59 \\
			\hline
			Baselines  
			& s-base   & -  & 62.25 \mypm 0.65 &  44.28 \mypm 0.63 & 70.84 \mypm 0.65 & 52.53 \mypm 0.59 \\
			
			& a-base    & \Checkmark &  64.98 \mypm 0.67 & 48.53 \mypm 0.64 & 67.96 \mypm 0.68 & 60.21 \mypm 0.69     \\
			
			& m-base    & \Checkmark &   78.08 \mypm 0.60 &  63.27 \mypm 0.70 &  75.90 \mypm 0.67 & 66.69 \mypm 0.68 \\
			\hline
			CD-FSL
			& m-base-FT~\cite{tseng2020cross}  & \Checkmark&   79.14  \mypm  0.62  &  65.42  \mypm  0.70  & 78.59  \mypm  0.60  &  68.26  \mypm  0.68 \\
			& meta-FDMixup (ours) & \Checkmark & \textbf{79.46 \mypm 0.63}  & \textbf{66.52 \mypm 0.70} & \textbf{78.92 \mypm 0.63} & \textbf{69.22 \mypm 0.65} \\
			\hline
		\end{tabular}
		\caption{The accuracy (\%) under the 5-way-1shot and 5-way-5-shot settings are reported. For all the baselines and CD-FSL methods, GNN is adopted as the FSL classifier. Our meta-FDMixup method performs best. }
		\label{tab:1}
	\end{table*}

	\begin{table*}
		\centering
		\begin{tabular}{lllll}
			\hline
			target set  &  mini-Imagnet $|$ CUB & mini-Imagenet $|$ Cars & mini-Imagnet $|$ Places & mini-Imagnet $|$ Plantae \\
			\hline
			
			s-base &   80.87 \mypm 0.56  & 80.87 \mypm 0.56  &  80.87 \mypm 0.56  & 80.87 \mypm 0.56 \\
			
			a-base & 57.83  \mypm 0.63  & 50.04  \mypm  0.59  & 64.15  \mypm  0.69  & 53.20  \mypm  0.63  \\
			
			m-base & 78.94  \mypm 0.58  & 80.75   \mypm 0.55  & 79.99   \mypm 0.58 & 80.51   \mypm 0.55 \\
			
			m-base-FT & 81.88 \mypm 0.57 & 80.89 \mypm 0.58 & 81.32   \mypm 0.56  & \textbf{82.28   \mypm 0.55} \\
			
			ours &  \textbf{82.29 \mypm 0.57}  & \textbf{81.00   \mypm 0.58} & \textbf{81.37   \mypm 0.56}   & 79.64 \mypm 0.59  \\
			\hline
			
		\end{tabular}
		\caption{Results (\%) on the testing set of  mini-ImagenetNet. The experiments are conducted under  5-way-5-shot setting taking GNN as  the FSL classifier. Our method outperforms the baselines and competitors in most cases. }
		\label{tab:5}
	\end{table*}

	\begin{table*}
		\centering
		\begin{tabular}{lllllll}
			\hline
			& \multicolumn{5}{c}{ single FSL classification task VS dual FSL classification tasks. }  \tabularnewline
			\hline
			setting & $D_{aux}$ & avg mini-ImageNet & CUB & Cars & Places & Plantae \\
			\hline
			$L_{FSL} = L_{FSL_S}$ &  \Checkmark  & 77.45 \mypm 0.63 & 46.08   \mypm 0.57     & 34.15   \mypm 0.53  & 65.41   \mypm 0.72   & 41.19   \mypm 0.60   \\
			
			$L_{FSL} = L_{FSL_A}$   &  \Checkmark  & 45.94 \mypm 0.58 &  74.18   \mypm 0.63   & 64.03   \mypm 0.71  & 67.51   \mypm 0.67     & 63.38   \mypm 0.69   \\
			
			meta-FDMixup        &  \Checkmark  &  \textbf{81.08 \mypm 0.58}  & \textbf{79.46 \mypm 0.63}  & \textbf{66.52 \mypm 0.70} & \textbf{78.92 \mypm 0.63} & \textbf{69.22 \mypm 0.65} \\ 
			\hline
			
			& \multicolumn{5}{c}{different strategies for $\lambda$}  \tabularnewline
			\hline
			setting & $D_{aux}$ & avg mini-Imagenet & CUB & Cars & Places & Plantae\\
			\hline
			$\lambda$-v1     &  \Checkmark  & 78.02 \mypm 0.60 &  78.21   \mypm 0.63   & 63.85   \mypm 0.70  & 76.51   \mypm 0.65  & 67.87   \mypm 0.72      \\
			
			$\lambda$-v2     & \Checkmark  & 80.78 \mypm 0.58 & 78.69   \mypm 0.63  & 60.65   \mypm 0.71   & 78.58   \mypm 0.63  & 66.91   \mypm 0.74  \\
			
			meta-FDMixup &  \Checkmark  & \textbf{81.08 \mypm 0.58}  & \textbf{79.46 \mypm 0.63}  & \textbf{66.52 \mypm 0.70} & \textbf{78.92 \mypm 0.63} & \textbf{69.22 \mypm 0.65} \\ 
			\hline
		\end{tabular}
		
		\caption{Ablation study on different FSL loss functions and different strategies for $\lambda$. The accuracy (\%) with GNN as the classifier under the 5-way-5-shot setting is reported. }
		\label{tab:4}
	\end{table*}

	\begin{figure*}
		\centering
		{\includegraphics[width=0.9\linewidth]{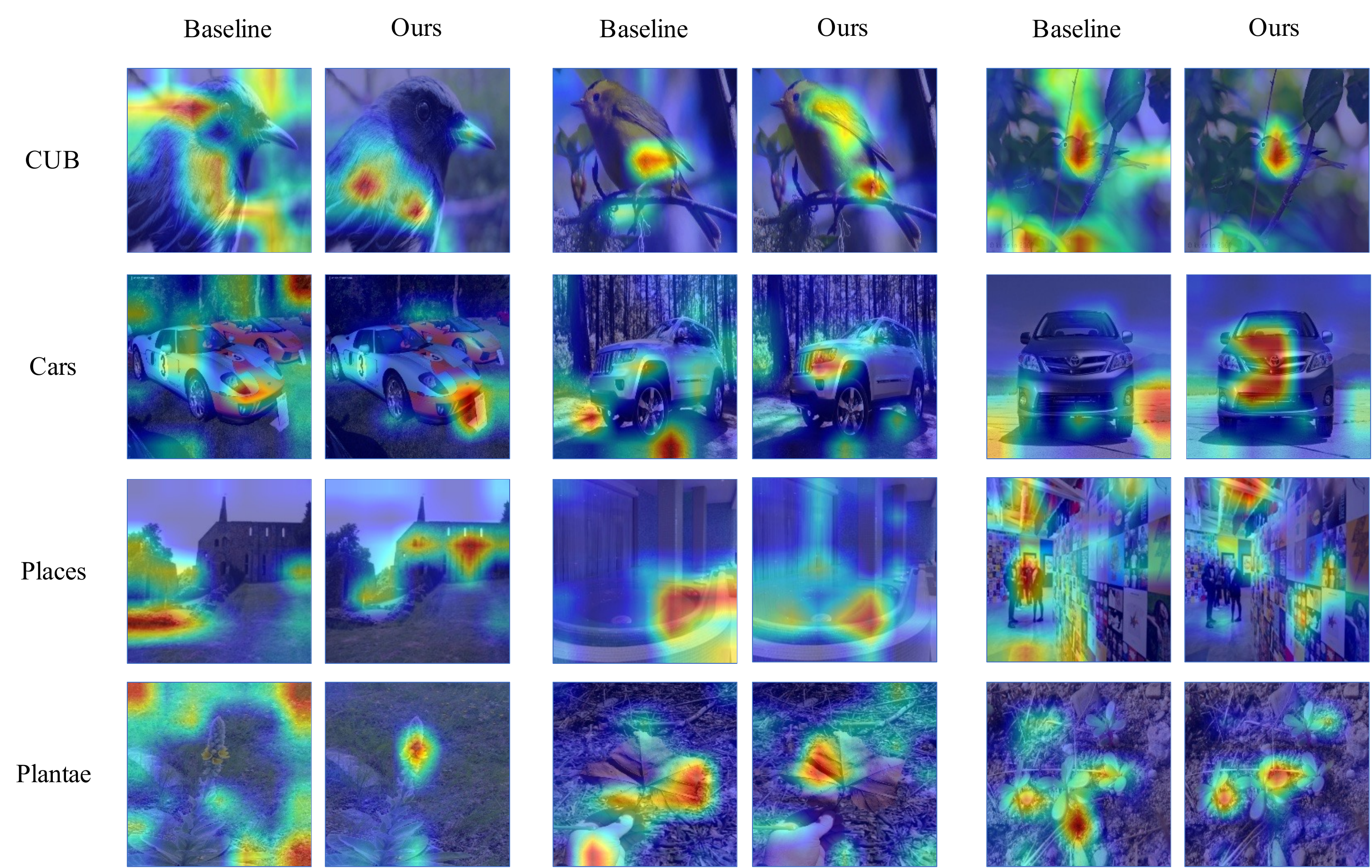}}
		\caption{Grad-CAM visualization results of the ``baseline" model and ``ours" model. For each target dataset, three examples are demonstrated. Our model meta-finetuned with the help of the auxiliary data can focus on the key areas of the target images.}
		\label{fig:vis} 
	\end{figure*}

	To show the effectiveness of our meta-FDMixup method, we compare it against all the baselines and competitors under the 5-way-1-shot and the 5-way-5-shot settings taking GNN as the FSL classifier. The results on the four target datasets are reported in Table~\ref{tab:1}.

	Besides, we also report 5-way-5-shot classification results on the mini-Imagenet dataset in Table~\ref{tab:5}. Since our trained model is target set specific, our results are reported in the form of ``mini-Imagenet $|$ target set". ``S-base" is the only exception, its results on the mini-Imagenet are not affected by the target set. 
	
	We highlight several points. 
	1) Comparing different FSL classifiers, the GNN performs best under this CD-FSL setting. It outperforms the Matching Net and the Relation Net by 10.88\% and 4.48\%  on the CUB dataset under the 5-way-5-shot setting, respectively. Thus, we take the GNN as our FSL classifier. 
	2) The ``s-base" baseline reaches a relatively good performance on the mini-Imagenet while fails to generalize to the target datasets with a sharp performance drop caused by the domain shift problem. 
	3) The ``a-base" baseline performs better than ``s-base" on the target datasets in most cases since it is meta-trained on the auxiliary dataset. However, its performance on the mini-Imagenet drops dramatically indicating that it suffers from the catastrophic forgetting problem. 
	4) Comparing with ``s-base", an obvious performance gain is brought by our strong baseline ``m-base" on the target datasets. Take the GNN classifier under 5-way-5-shot as an example, we improve the accuracy by 15.83\%, 18.99\%, 5.06\%, and 14.16\% on CUB, Cars, Places, and Plantae datasets, respectively. We show that our setting brings very considerable benefits with only extremely few labeled data introduced. 
	5) Another observation worth mentioning about the ``m-base" is that though it performs well on the target dataset, the ability to recognize the initial mini-Imagenet is damaged. 
	That is because merging the relatively large-scale mini-Imagenet with the small-scale auxiliary dataset may cause the data imbalance problem.
	Not to mention, they belong to different domains. 
	6) For the competitor ``m-base-FT", we notice that it steadily improve the performance of the model on both the source and the target datasets which indicates its effectiveness. However, manually fine-tuning the hyper-parameters needs a lot of human labor, while our method is learned automatically.
	7) Our method outperforms all the above methods on the source dataset and the different target datasets. Under the 5-way-5-shot setting, our meta-FDMixup achieves 79.46\%, 66.52\%, 78.92\%, 69.22\% on CUB, Cars, Places, and Plantae, respectively. Most importantly, the aforementioned problems encountered when using the small-regime auxiliary dataset are tackled by our meta-FDMixup.
	
	Our mixup module enables us to mix the images from different domains resulting in massive and diverse mixed images.  Our disentangle module learns to decouple the domain-irrelevant feature and the domain-specific feature, thus avoiding the influence of the domain shift. 
	As a result, we not only achieve the best performance on the target datasets but also improve the accuracy on the mini-Imagenet dataset, which is actually out of our expectations.

	
	\subsection{Ablation Study}
	We first explore the effectiveness of our dual FSL classification task learning mechanism, then study the impact of $\lambda$ on our meta-FDMixup method. We conduct the experiments taking GNN as the classifier under the 5-way-5-shot learning setting. The results are demonstrated in Table~\ref{tab:4}. Notably, due to the limitation of the manuscript, the average accuracy on mini-ImageNet given different target datasets is reported.

	\noindent\textbf{Single FSL classification task VS dual FSL classification tasks. } We compare our mechanism of dual few-shot classification tasks with that of single task. Specifically, to shown the performance of single FSL task, we define our FSL classification loss function as ``$L_{FSL} = L_{FSL_S}$ " and ``$L_{FSL} = L_{FSL_{A}}$ ", respectively. 
	The results are shown in the first part of Table~\ref{tab:4}. An interesting phenomenon is that the single task mechanism is inferior to our multi-task mechanism even on its goal dataset. This observation in turn indicates that our multi-task mechanism makes the source and the target datasets promote each other mutually.

	\noindent\textbf{Different strategies for $\lambda$.} In our meta-FDMixup, the $\lambda \sim Beta(\alpha, \alpha)$ treating the mini-Imagenet and the auxiliary data equally. We are wondering what would happen if we assign a higher ratio to one of these two datasets. In the end, two simple experiments are conducted. For ``$\lambda$-v1", if the value of $\lambda$ is higher than 0.5, we adjust it to 0.5, which means we want to ensure more than half of the auxiliary data will be maintained in the mixed data. Similarly, for ``$\lambda$-v2", we set the $\lambda$ to 0.5 if it is smaller than 0.5. 
	The results are shown in the second part of Table~\ref{tab:4}. We notice that the accuracy of the target set decreases in both cases. For ``$\lambda$-v2" it is easy to understand since it tends to protect the mini-Imagenet data. One reason that causes the performance drop for the ``$\lambda$-v1" may be that the diversity of the mixed data is also reduced with a smaller ratio of mini-Imagenet.

	\subsection{Visualization Results}
	To provide a more intuitive understanding of our new setting -- introducing some auxiliary target images to the training stage, we visualize the class-activation map of ``baseline" and ``ours" using the grad-CAM~\cite{selvaraju2017grad}. ``Baseline" refers to the model pretrained on the mini-ImageNet dataset. It is composed of the feature encoder, the disentangle module, and the FC classifier. ``Ours" denotes our meta-FDMixup model further meta-finetuned on the source dataset and the auxiliary dataset. For each target dataset, three examples are sampled from the testing set to give a comparison between the ``baseline" and the ``ours".  The results are shown in Figure~\ref{fig:vis}.

	Several phenomenons can be observed. 1) Generally, the ``baseline" model pays more attention to the things which are more common in the mini-Imagenet, such as the branches in the CUB dataset, the mountains in the Cars dataset, and some backgrounds in the Places and the Plantae datasets. This will damage the model performance in the target datasets.  However, without any target dataset, it is difficult to generalize to the target datasets well.  2) ``Ours" model alleviates this problem to some extent. We successfully move the model's attention to those areas that are critical for identifying the target classes. Take the CUB dataset as an example, our model focuses on the key parts of the birds, such as the feathers and the tails.

	\section{Conclusion}
	In this paper, to tackle the domain shift problem in CD-FSL, we advocate utilizing extremely few labeled target data for cross-domain few-shot learning. The detailed pilot and feasibility and studies explore the most suitable training strategy and validate the soundness of our new setting. Technically, a novel meta-FDMixup network that collaborates with meta-learning is proposed by us. By mixing the images of the source and the target datasets and learning to disentangle the domain-irrelevant and domain-specific features, our meta-FDMixup model can be well generalized to the target datasets. Extensive experiments show the effectiveness of our method.

	\section{Acknowledgement}
	This work was supported by National Natural Science Foundation of China Project (62072116) and Science and Technology Commission of Shanghai Municipality Projects (19511120700, 19ZR1471800).

	\bibliographystyle{ACM-Reference-Format}
	\balance
	\bibliography{references}


\begin{thebibliography}{38}


\ifx \showCODEN    \undefined \def \showCODEN     #1{\unskip}     \fi
\ifx \showDOI      \undefined \def \showDOI       #1{#1}\fi
\ifx \showISBNx    \undefined \def \showISBNx     #1{\unskip}     \fi
\ifx \showISBNxiii \undefined \def \showISBNxiii  #1{\unskip}     \fi
\ifx \showISSN     \undefined \def \showISSN      #1{\unskip}     \fi
\ifx \showLCCN     \undefined \def \showLCCN      #1{\unskip}     \fi
\ifx \shownote     \undefined \def \shownote      #1{#1}          \fi
\ifx \showarticletitle \undefined \def \showarticletitle #1{#1}   \fi
\ifx \showURL      \undefined \def \showURL       {\relax}        \fi
\providecommand\bibfield[2]{#2}
\providecommand\bibinfo[2]{#2}
\providecommand\natexlab[1]{#1}
\providecommand\showeprint[2][]{arXiv:#2}

\bibitem[\protect\citeauthoryear{Bousmalis, Silberman, Dohan, Erhan, and
  Krishnan}{Bousmalis et~al\mbox{.}}{2017}]%
        {bousmalis2017unsupervised}
\bibfield{author}{\bibinfo{person}{Konstantinos Bousmalis},
  \bibinfo{person}{Nathan Silberman}, \bibinfo{person}{David Dohan},
  \bibinfo{person}{Dumitru Erhan}, {and} \bibinfo{person}{Dilip Krishnan}.}
  \bibinfo{year}{2017}\natexlab{}.
\newblock \showarticletitle{Unsupervised pixel-level domain adaptation with
  generative adversarial networks}. In \bibinfo{booktitle}{\emph{CVPR}}.
\newblock


\bibitem[\protect\citeauthoryear{Bousmalis, Trigeorgis, Silberman, Krishnan,
  and Erhan}{Bousmalis et~al\mbox{.}}{2016}]%
        {bousmalis2016domain}
\bibfield{author}{\bibinfo{person}{Konstantinos Bousmalis},
  \bibinfo{person}{George Trigeorgis}, \bibinfo{person}{Nathan Silberman},
  \bibinfo{person}{Dilip Krishnan}, {and} \bibinfo{person}{Dumitru Erhan}.}
  \bibinfo{year}{2016}\natexlab{}.
\newblock \showarticletitle{Domain separation networks}.
\newblock \bibinfo{journal}{\emph{arXiv preprint}} (\bibinfo{year}{2016}).
\newblock


\bibitem[\protect\citeauthoryear{Cai, Cai, and Shen}{Cai et~al\mbox{.}}{2020}]%
        {cai2020sb}
\bibfield{author}{\bibinfo{person}{John Cai}, \bibinfo{person}{Bill Cai}, {and}
  \bibinfo{person}{Sheng~Mei Shen}.} \bibinfo{year}{2020}\natexlab{}.
\newblock \showarticletitle{SB-MTL: Score-based Meta Transfer-Learning for
  Cross-Domain Few-Shot Learning}.
\newblock \bibinfo{journal}{\emph{arXiv preprint arXiv:2012.01784}}
  (\bibinfo{year}{2020}).
\newblock


\bibitem[\protect\citeauthoryear{Chen, Liu, Kira, Wang, and Huang}{Chen
  et~al\mbox{.}}{2019b}]%
        {chen2019closer}
\bibfield{author}{\bibinfo{person}{Wei-Yu Chen}, \bibinfo{person}{Yen-Cheng
  Liu}, \bibinfo{person}{Zsolt Kira}, \bibinfo{person}{Yu-Chiang~Frank Wang},
  {and} \bibinfo{person}{Jia-Bin Huang}.} \bibinfo{year}{2019}\natexlab{b}.
\newblock \showarticletitle{A closer look at few-shot classification}.
\newblock \bibinfo{journal}{\emph{arXiv preprint}} (\bibinfo{year}{2019}).
\newblock


\bibitem[\protect\citeauthoryear{Chen, Fu, Chen, and Jiang}{Chen
  et~al\mbox{.}}{2019a}]%
        {chen2019image}
\bibfield{author}{\bibinfo{person}{Zitian Chen}, \bibinfo{person}{Yanwei Fu},
  \bibinfo{person}{Kaiyu Chen}, {and} \bibinfo{person}{Yu-Gang Jiang}.}
  \bibinfo{year}{2019}\natexlab{a}.
\newblock \showarticletitle{Image block augmentation for one-shot learning}. In
  \bibinfo{booktitle}{\emph{AAAI}}.
\newblock


\bibitem[\protect\citeauthoryear{Fei, Lu, Xiang, and Huang}{Fei
  et~al\mbox{.}}{[n.d.]}]%
        {feimelr}
\bibfield{author}{\bibinfo{person}{Nanyi Fei}, \bibinfo{person}{Zhiwu Lu},
  \bibinfo{person}{Tao Xiang}, {and} \bibinfo{person}{Songfang Huang}.}
  \bibinfo{year}{[n.d.]}\natexlab{}.
\newblock \showarticletitle{MELR: Meta-Learning via Modeling Episode-level
  Relationships for Few-shot Learning}.
\newblock  (\bibinfo{year}{[n.\,d.]}).
\newblock


\bibitem[\protect\citeauthoryear{Finn, Abbeel, and Levine}{Finn
  et~al\mbox{.}}{2017}]%
        {finn2017model}
\bibfield{author}{\bibinfo{person}{Chelsea Finn}, \bibinfo{person}{Pieter
  Abbeel}, {and} \bibinfo{person}{Sergey Levine}.}
  \bibinfo{year}{2017}\natexlab{}.
\newblock \showarticletitle{Model-agnostic meta-learning for fast adaptation of
  deep networks}.
\newblock \bibinfo{journal}{\emph{arXiv preprint}} (\bibinfo{year}{2017}).
\newblock


\bibitem[\protect\citeauthoryear{Ganin, Ustinova, Ajakan, Germain, Larochelle,
  Laviolette, Marchand, and Lempitsky}{Ganin et~al\mbox{.}}{2016}]%
        {ganin2016domain}
\bibfield{author}{\bibinfo{person}{Yaroslav Ganin}, \bibinfo{person}{Evgeniya
  Ustinova}, \bibinfo{person}{Hana Ajakan}, \bibinfo{person}{Pascal Germain},
  \bibinfo{person}{Hugo Larochelle}, \bibinfo{person}{Fran{\c{c}}ois
  Laviolette}, \bibinfo{person}{Mario Marchand}, {and} \bibinfo{person}{Victor
  Lempitsky}.} \bibinfo{year}{2016}\natexlab{}.
\newblock \showarticletitle{Domain-adversarial training of neural networks}.
\newblock \bibinfo{journal}{\emph{JMLR}} (\bibinfo{year}{2016}).
\newblock


\bibitem[\protect\citeauthoryear{Garcia and Bruna}{Garcia and Bruna}{2017}]%
        {garcia2017few}
\bibfield{author}{\bibinfo{person}{Victor Garcia} {and} \bibinfo{person}{Joan
  Bruna}.} \bibinfo{year}{2017}\natexlab{}.
\newblock \showarticletitle{Few-shot learning with graph neural networks}.
\newblock \bibinfo{journal}{\emph{arXiv preprint}} (\bibinfo{year}{2017}).
\newblock


\bibitem[\protect\citeauthoryear{Ghifary, Kleijn, Zhang, Balduzzi, and
  Li}{Ghifary et~al\mbox{.}}{2016}]%
        {ghifary2016deep}
\bibfield{author}{\bibinfo{person}{Muhammad Ghifary},
  \bibinfo{person}{W~Bastiaan Kleijn}, \bibinfo{person}{Mengjie Zhang},
  \bibinfo{person}{David Balduzzi}, {and} \bibinfo{person}{Wen Li}.}
  \bibinfo{year}{2016}\natexlab{}.
\newblock \showarticletitle{Deep reconstruction-classification networks for
  unsupervised domain adaptation}. In \bibinfo{booktitle}{\emph{ECCV}}.
\newblock


\bibitem[\protect\citeauthoryear{Guo, Codella, Karlinsky, Codella, Smith,
  Saenko, Rosing, and Feris}{Guo et~al\mbox{.}}{2020}]%
        {guo2020broader}
\bibfield{author}{\bibinfo{person}{Yunhui Guo}, \bibinfo{person}{Noel~C
  Codella}, \bibinfo{person}{Leonid Karlinsky}, \bibinfo{person}{James~V
  Codella}, \bibinfo{person}{John~R Smith}, \bibinfo{person}{Kate Saenko},
  \bibinfo{person}{Tajana Rosing}, {and} \bibinfo{person}{Rogerio Feris}.}
  \bibinfo{year}{2020}\natexlab{}.
\newblock \showarticletitle{A broader study of cross-domain few-shot learning}.
  In \bibinfo{booktitle}{\emph{ECCV}}.
\newblock


\bibitem[\protect\citeauthoryear{He, Zhang, Ren, and Sun}{He
  et~al\mbox{.}}{2016}]%
        {he2016deep}
\bibfield{author}{\bibinfo{person}{Kaiming He}, \bibinfo{person}{Xiangyu
  Zhang}, \bibinfo{person}{Shaoqing Ren}, {and} \bibinfo{person}{Jian Sun}.}
  \bibinfo{year}{2016}\natexlab{}.
\newblock \showarticletitle{Deep residual learning for image recognition}. In
  \bibinfo{booktitle}{\emph{CVPR}}.
\newblock


\bibitem[\protect\citeauthoryear{Hendrycks, Mu, Cubuk, Zoph, Gilmer, and
  Lakshminarayanan}{Hendrycks et~al\mbox{.}}{2019}]%
        {hendrycks2019augmix}
\bibfield{author}{\bibinfo{person}{Dan Hendrycks}, \bibinfo{person}{Norman Mu},
  \bibinfo{person}{Ekin~D Cubuk}, \bibinfo{person}{Barret Zoph},
  \bibinfo{person}{Justin Gilmer}, {and} \bibinfo{person}{Balaji
  Lakshminarayanan}.} \bibinfo{year}{2019}\natexlab{}.
\newblock \showarticletitle{Augmix: A simple data processing method to improve
  robustness and uncertainty}.
\newblock \bibinfo{journal}{\emph{arXiv preprint arXiv:1912.02781}}
  (\bibinfo{year}{2019}).
\newblock


\bibitem[\protect\citeauthoryear{Hou, Chang, Ma, Shan, and Chen}{Hou
  et~al\mbox{.}}{2019}]%
        {hou2019cross}
\bibfield{author}{\bibinfo{person}{Ruibing Hou}, \bibinfo{person}{Hong Chang},
  \bibinfo{person}{Bingpeng Ma}, \bibinfo{person}{Shiguang Shan}, {and}
  \bibinfo{person}{Xilin Chen}.} \bibinfo{year}{2019}\natexlab{}.
\newblock \showarticletitle{Cross attention network for few-shot
  classification}.
\newblock \bibinfo{journal}{\emph{arXiv preprint arXiv:1910.07677}}
  (\bibinfo{year}{2019}).
\newblock


\bibitem[\protect\citeauthoryear{Kang, Jiang, Yang, and Hauptmann}{Kang
  et~al\mbox{.}}{2019}]%
        {kang2019contrastive}
\bibfield{author}{\bibinfo{person}{Guoliang Kang}, \bibinfo{person}{Lu Jiang},
  \bibinfo{person}{Yi Yang}, {and} \bibinfo{person}{Alexander~G Hauptmann}.}
  \bibinfo{year}{2019}\natexlab{}.
\newblock \showarticletitle{Contrastive adaptation network for unsupervised
  domain adaptation}. In \bibinfo{booktitle}{\emph{CVPR}}.
\newblock


\bibitem[\protect\citeauthoryear{Kim, Choo, and Song}{Kim
  et~al\mbox{.}}{2020}]%
        {kim2020puzzle}
\bibfield{author}{\bibinfo{person}{Jang-Hyun Kim}, \bibinfo{person}{Wonho
  Choo}, {and} \bibinfo{person}{Hyun~Oh Song}.}
  \bibinfo{year}{2020}\natexlab{}.
\newblock \showarticletitle{Puzzle mix: Exploiting saliency and local
  statistics for optimal mixup}. In \bibinfo{booktitle}{\emph{International
  Conference on Machine Learning}}. PMLR, \bibinfo{pages}{5275--5285}.
\newblock


\bibitem[\protect\citeauthoryear{Krause, Stark, Deng, and Fei-Fei}{Krause
  et~al\mbox{.}}{2013}]%
        {krause20133d}
\bibfield{author}{\bibinfo{person}{Jonathan Krause}, \bibinfo{person}{Michael
  Stark}, \bibinfo{person}{Jia Deng}, {and} \bibinfo{person}{Li Fei-Fei}.}
  \bibinfo{year}{2013}\natexlab{}.
\newblock \showarticletitle{3d object representations for fine-grained
  categorization}. In \bibinfo{booktitle}{\emph{ICCVW}}.
\newblock


\bibitem[\protect\citeauthoryear{Li, Zhang, Li, and Fu}{Li
  et~al\mbox{.}}{2020b}]%
        {li2020adversarial}
\bibfield{author}{\bibinfo{person}{Kai Li}, \bibinfo{person}{Yulun Zhang},
  \bibinfo{person}{Kunpeng Li}, {and} \bibinfo{person}{Yun Fu}.}
  \bibinfo{year}{2020}\natexlab{b}.
\newblock \showarticletitle{Adversarial Feature Hallucination Networks for
  Few-Shot Learning}. In \bibinfo{booktitle}{\emph{CVPR}}.
\newblock


\bibitem[\protect\citeauthoryear{Li, Xiong, An, Xu, and Dou}{Li
  et~al\mbox{.}}{2020a}]%
        {li2020xmixup}
\bibfield{author}{\bibinfo{person}{Xingjian Li}, \bibinfo{person}{Haoyi Xiong},
  \bibinfo{person}{Haozhe An}, \bibinfo{person}{Chengzhong Xu}, {and}
  \bibinfo{person}{Dejing Dou}.} \bibinfo{year}{2020}\natexlab{a}.
\newblock \showarticletitle{XMixup: Efficient Transfer Learning with Auxiliary
  Samples by Cross-domain Mixup}.
\newblock \bibinfo{journal}{\emph{arXiv preprint}} (\bibinfo{year}{2020}).
\newblock


\bibitem[\protect\citeauthoryear{Phoo and Hariharan}{Phoo and
  Hariharan}{2020}]%
        {phoo2020self}
\bibfield{author}{\bibinfo{person}{Cheng~Perng Phoo} {and}
  \bibinfo{person}{Bharath Hariharan}.} \bibinfo{year}{2020}\natexlab{}.
\newblock \showarticletitle{Self-training for Few-shot Transfer Across Extreme
  Task Differences}.
\newblock \bibinfo{journal}{\emph{arXiv preprint}} (\bibinfo{year}{2020}).
\newblock


\bibitem[\protect\citeauthoryear{Ravi and Larochelle}{Ravi and
  Larochelle}{2017}]%
        {ravi2016optimization}
\bibfield{author}{\bibinfo{person}{Sachin Ravi} {and} \bibinfo{person}{Hugo
  Larochelle}.} \bibinfo{year}{2017}\natexlab{}.
\newblock \showarticletitle{Optimization as a model for few-shot learning}. In
  \bibinfo{booktitle}{\emph{ICLR}}.
\newblock


\bibitem[\protect\citeauthoryear{Rozantsev, Salzmann, and Fua}{Rozantsev
  et~al\mbox{.}}{2018}]%
        {rozantsev2018beyond}
\bibfield{author}{\bibinfo{person}{Artem Rozantsev}, \bibinfo{person}{Mathieu
  Salzmann}, {and} \bibinfo{person}{Pascal Fua}.}
  \bibinfo{year}{2018}\natexlab{}.
\newblock \showarticletitle{Beyond sharing weights for deep domain adaptation}.
\newblock \bibinfo{journal}{\emph{TPAMI}} (\bibinfo{year}{2018}).
\newblock


\bibitem[\protect\citeauthoryear{Rusu, Rao, Sygnowski, Vinyals, Pascanu,
  Osindero, and Hadsell}{Rusu et~al\mbox{.}}{2018}]%
        {rusu2018meta}
\bibfield{author}{\bibinfo{person}{Andrei~A Rusu}, \bibinfo{person}{Dushyant
  Rao}, \bibinfo{person}{Jakub Sygnowski}, \bibinfo{person}{Oriol Vinyals},
  \bibinfo{person}{Razvan Pascanu}, \bibinfo{person}{Simon Osindero}, {and}
  \bibinfo{person}{Raia Hadsell}.} \bibinfo{year}{2018}\natexlab{}.
\newblock \showarticletitle{Meta-learning with latent embedding optimization}.
\newblock \bibinfo{journal}{\emph{arXiv preprint}} (\bibinfo{year}{2018}).
\newblock


\bibitem[\protect\citeauthoryear{Selvaraju, Cogswell, Das, Vedantam, Parikh,
  and Batra}{Selvaraju et~al\mbox{.}}{2017}]%
        {selvaraju2017grad}
\bibfield{author}{\bibinfo{person}{Ramprasaath~R Selvaraju},
  \bibinfo{person}{Michael Cogswell}, \bibinfo{person}{Abhishek Das},
  \bibinfo{person}{Ramakrishna Vedantam}, \bibinfo{person}{Devi Parikh}, {and}
  \bibinfo{person}{Dhruv Batra}.} \bibinfo{year}{2017}\natexlab{}.
\newblock \showarticletitle{Grad-cam: Visual explanations from deep networks
  via gradient-based localization}. In \bibinfo{booktitle}{\emph{Proceedings of
  the IEEE international conference on computer vision}}.
  \bibinfo{pages}{618--626}.
\newblock


\bibitem[\protect\citeauthoryear{Snell, Swersky, and Zemel}{Snell
  et~al\mbox{.}}{2017}]%
        {snell2017prototypical}
\bibfield{author}{\bibinfo{person}{Jake Snell}, \bibinfo{person}{Kevin
  Swersky}, {and} \bibinfo{person}{Richard Zemel}.}
  \bibinfo{year}{2017}\natexlab{}.
\newblock \showarticletitle{Prototypical networks for few-shot learning}. In
  \bibinfo{booktitle}{\emph{NeurIPS}}.
\newblock


\bibitem[\protect\citeauthoryear{Sun, Lapuschkin, Samek, Zhao, Cheung, and
  Binder}{Sun et~al\mbox{.}}{2020}]%
        {sun2020explanation}
\bibfield{author}{\bibinfo{person}{Jiamei Sun}, \bibinfo{person}{Sebastian
  Lapuschkin}, \bibinfo{person}{Wojciech Samek}, \bibinfo{person}{Yunqing
  Zhao}, \bibinfo{person}{Ngai-Man Cheung}, {and} \bibinfo{person}{Alexander
  Binder}.} \bibinfo{year}{2020}\natexlab{}.
\newblock \showarticletitle{Explanation-guided training for cross-domain
  few-shot classification}.
\newblock \bibinfo{journal}{\emph{arXiv preprint}} (\bibinfo{year}{2020}).
\newblock


\bibitem[\protect\citeauthoryear{Sung, Yang, Zhang, Xiang, Torr, and
  Hospedales}{Sung et~al\mbox{.}}{2018}]%
        {sung2018learning}
\bibfield{author}{\bibinfo{person}{Flood Sung}, \bibinfo{person}{Yongxin Yang},
  \bibinfo{person}{Li Zhang}, \bibinfo{person}{Tao Xiang},
  \bibinfo{person}{Philip~HS Torr}, {and} \bibinfo{person}{Timothy~M
  Hospedales}.} \bibinfo{year}{2018}\natexlab{}.
\newblock \showarticletitle{Learning to compare: Relation network for few-shot
  learning}. In \bibinfo{booktitle}{\emph{CVPR}}.
\newblock


\bibitem[\protect\citeauthoryear{Tseng, Lee, Huang, and Yang}{Tseng
  et~al\mbox{.}}{2020}]%
        {tseng2020cross}
\bibfield{author}{\bibinfo{person}{Hung-Yu Tseng}, \bibinfo{person}{Hsin-Ying
  Lee}, \bibinfo{person}{Jia-Bin Huang}, {and} \bibinfo{person}{Ming-Hsuan
  Yang}.} \bibinfo{year}{2020}\natexlab{}.
\newblock \showarticletitle{Cross-domain few-shot classification via learned
  feature-wise transformation}. In \bibinfo{booktitle}{\emph{ICLR}}.
\newblock


\bibitem[\protect\citeauthoryear{Tzeng, Hoffman, Darrell, and Saenko}{Tzeng
  et~al\mbox{.}}{2015}]%
        {tzeng2015simultaneous}
\bibfield{author}{\bibinfo{person}{Eric Tzeng}, \bibinfo{person}{Judy Hoffman},
  \bibinfo{person}{Trevor Darrell}, {and} \bibinfo{person}{Kate Saenko}.}
  \bibinfo{year}{2015}\natexlab{}.
\newblock \showarticletitle{Simultaneous deep transfer across domains and
  tasks}. In \bibinfo{booktitle}{\emph{ICCV}}.
\newblock


\bibitem[\protect\citeauthoryear{Tzeng, Hoffman, Zhang, Saenko, and
  Darrell}{Tzeng et~al\mbox{.}}{2014}]%
        {tzeng2014deep}
\bibfield{author}{\bibinfo{person}{Eric Tzeng}, \bibinfo{person}{Judy Hoffman},
  \bibinfo{person}{Ning Zhang}, \bibinfo{person}{Kate Saenko}, {and}
  \bibinfo{person}{Trevor Darrell}.} \bibinfo{year}{2014}\natexlab{}.
\newblock \showarticletitle{Deep domain confusion: Maximizing for domain
  invariance}.
\newblock \bibinfo{journal}{\emph{arXiv preprint}} (\bibinfo{year}{2014}).
\newblock


\bibitem[\protect\citeauthoryear{Van~Horn, Mac~Aodha, Song, Cui, Sun, Shepard,
  Adam, Perona, and Belongie}{Van~Horn et~al\mbox{.}}{2018}]%
        {van2018inaturalist}
\bibfield{author}{\bibinfo{person}{Grant Van~Horn}, \bibinfo{person}{Oisin
  Mac~Aodha}, \bibinfo{person}{Yang Song}, \bibinfo{person}{Yin Cui},
  \bibinfo{person}{Chen Sun}, \bibinfo{person}{Alex Shepard},
  \bibinfo{person}{Hartwig Adam}, \bibinfo{person}{Pietro Perona}, {and}
  \bibinfo{person}{Serge Belongie}.} \bibinfo{year}{2018}\natexlab{}.
\newblock \showarticletitle{The inaturalist species classification and
  detection dataset}. In \bibinfo{booktitle}{\emph{CVPR}}.
\newblock


\bibitem[\protect\citeauthoryear{Verma, Lamb, Beckham, Najafi, Mitliagkas,
  Lopez-Paz, and Bengio}{Verma et~al\mbox{.}}{2019}]%
        {verma2019manifold}
\bibfield{author}{\bibinfo{person}{Vikas Verma}, \bibinfo{person}{Alex Lamb},
  \bibinfo{person}{Christopher Beckham}, \bibinfo{person}{Amir Najafi},
  \bibinfo{person}{Ioannis Mitliagkas}, \bibinfo{person}{David Lopez-Paz},
  {and} \bibinfo{person}{Yoshua Bengio}.} \bibinfo{year}{2019}\natexlab{}.
\newblock \showarticletitle{Manifold mixup: Better representations by
  interpolating hidden states}. In \bibinfo{booktitle}{\emph{ICML}}.
\newblock


\bibitem[\protect\citeauthoryear{Vinyals, Blundell, Lillicrap, Wierstra,
  et~al\mbox{.}}{Vinyals et~al\mbox{.}}{2016}]%
        {vinyals2016matching}
\bibfield{author}{\bibinfo{person}{Oriol Vinyals}, \bibinfo{person}{Charles
  Blundell}, \bibinfo{person}{Timothy Lillicrap}, \bibinfo{person}{Daan
  Wierstra}, {et~al\mbox{.}}} \bibinfo{year}{2016}\natexlab{}.
\newblock \showarticletitle{Matching networks for one shot learning}. In
  \bibinfo{booktitle}{\emph{NeurIPS}}.
\newblock


\bibitem[\protect\citeauthoryear{Wah, Branson, Welinder, Perona, and
  Belongie}{Wah et~al\mbox{.}}{2011}]%
        {wah2011caltech}
\bibfield{author}{\bibinfo{person}{Catherine Wah}, \bibinfo{person}{Steve
  Branson}, \bibinfo{person}{Peter Welinder}, \bibinfo{person}{Pietro Perona},
  {and} \bibinfo{person}{Serge Belongie}.} \bibinfo{year}{2011}\natexlab{}.
\newblock \showarticletitle{The caltech-ucsd birds-200-2011 dataset}.
\newblock  (\bibinfo{year}{2011}).
\newblock


\bibitem[\protect\citeauthoryear{Ye, Hu, Zhan, and Sha}{Ye
  et~al\mbox{.}}{2020}]%
        {ye2020few}
\bibfield{author}{\bibinfo{person}{Han-Jia Ye}, \bibinfo{person}{Hexiang Hu},
  \bibinfo{person}{De-Chuan Zhan}, {and} \bibinfo{person}{Fei Sha}.}
  \bibinfo{year}{2020}\natexlab{}.
\newblock \showarticletitle{Few-shot learning via embedding adaptation with
  set-to-set functions}. In \bibinfo{booktitle}{\emph{Proceedings of the
  IEEE/CVF Conference on Computer Vision and Pattern Recognition}}.
  \bibinfo{pages}{8808--8817}.
\newblock


\bibitem[\protect\citeauthoryear{Yun, Han, Oh, Chun, Choe, and Yoo}{Yun
  et~al\mbox{.}}{2019}]%
        {yun2019cutmix}
\bibfield{author}{\bibinfo{person}{Sangdoo Yun}, \bibinfo{person}{Dongyoon
  Han}, \bibinfo{person}{Seong~Joon Oh}, \bibinfo{person}{Sanghyuk Chun},
  \bibinfo{person}{Junsuk Choe}, {and} \bibinfo{person}{Youngjoon Yoo}.}
  \bibinfo{year}{2019}\natexlab{}.
\newblock \showarticletitle{Cutmix: Regularization strategy to train strong
  classifiers with localizable features}. In \bibinfo{booktitle}{\emph{ICCV}}.
\newblock


\bibitem[\protect\citeauthoryear{Zhang, Cisse, Dauphin, and Lopez-Paz}{Zhang
  et~al\mbox{.}}{2017}]%
        {zhang2017mixup}
\bibfield{author}{\bibinfo{person}{Hongyi Zhang}, \bibinfo{person}{Moustapha
  Cisse}, \bibinfo{person}{Yann~N Dauphin}, {and} \bibinfo{person}{David
  Lopez-Paz}.} \bibinfo{year}{2017}\natexlab{}.
\newblock \showarticletitle{mixup: Beyond empirical risk minimization}.
\newblock \bibinfo{journal}{\emph{arXiv preprint}} (\bibinfo{year}{2017}).
\newblock


\bibitem[\protect\citeauthoryear{Zhou, Lapedriza, Khosla, Oliva, and
  Torralba}{Zhou et~al\mbox{.}}{2017}]%
        {zhou2017places}
\bibfield{author}{\bibinfo{person}{Bolei Zhou}, \bibinfo{person}{Agata
  Lapedriza}, \bibinfo{person}{Aditya Khosla}, \bibinfo{person}{Aude Oliva},
  {and} \bibinfo{person}{Antonio Torralba}.} \bibinfo{year}{2017}\natexlab{}.
\newblock \showarticletitle{Places: A 10 million image database for scene
  recognition}.
\newblock \bibinfo{journal}{\emph{TPAMI}} (\bibinfo{year}{2017}).
\newblock


\end{thebibliography}

\end{document}